\documentclass{article}

\usepackage{bbm}
\usepackage{amsmath}
\usepackage{graphicx}
\usepackage{subcaption}
\usepackage[preprint]{neurips_2025}

\usepackage[utf8]{inputenc} 
\usepackage[T1]{fontenc}    
\usepackage{hyperref}       
\usepackage{url}            
\usepackage{booktabs}       
\usepackage{amsfonts}       
\usepackage{nicefrac}       
\usepackage{microtype}      
\usepackage{xcolor}         

\title{Position: There Is No Free Bayesian Uncertainty Quantification}

\author{%
  Ivan Melev\thanks{corresponding author}\\
  Department of Statistics\\
  LMU Munich\\
  Ludwigst. 33, 80539, Munich \\
  \texttt{ivan.melev@lmu.de} \\
  \And
  Goeran Kauermann \\
  Department of Statistics\\
  LMU Munich\\
  Ludwigst. 33, 80539, Munich \\
   \texttt{goeran.kauermann@stat.uni-muenchen.de} \\
}

\begin{document}

\maketitle

\begin{abstract}
  Due to their intuitive appeal, Bayesian methods of modeling and uncertainty quantification have become popular in modern machine and deep learning.
  When providing a prior distribution over the parameter space, it is straightforward to obtain a distribution over the parameters that is conventionally interpreted as uncertainty quantification of the model.
  We challenge the validity of such Bayesian uncertainty quantification by discussing the equivalent optimization-based representation of Bayesian updating, provide an alternative interpretation that is coherent with the optimization-based perspective, propose measures of the quality of the Bayesian inferential stage, and suggest directions for future work.
\end{abstract}

\section{Introduction}
Machine Learning (ML) and statistical learning rely on data, however the data is always finite and for some application areas rather limited.
This inevitably leads to uncertainty in the estimation, representation or prediction stages, which needs to be quantified.
Although these three types of uncertainty are different, they are addressed by the same set of techniques. 
Multiple types of uncertainty quantification (UQ) (for a review check \cite{hullermeier2021aleatoric}) have been developed; however, they can be coarsely categorized as belonging either to the frequentist or Bayesian framework.
\paragraph{Frequentist UQ}
The frequentist framework treats the data as random variables and the parameters as fixed but unknown.
Therefore, it relies on the construction of estimators that are treated as random functions of the data. 
Consequently, the uncertainty arises from the stochasticity in the data, with a particular dataset being just one realization out of potentially uncountably many (\cite{wasserman2013all, samaniego2010comparison}).
Frequentist UQ does not inform the modeler's beliefs; however, its reliance on the distribution of the chosen estimator leads to desirable long run properties regarding coverage of the fixed but unknown parameters, as well as valid predictive coverage.
\paragraph{Bayesian UQ}
The Bayesian framework treats the parameters as random variables and the data as fixed, providing the information on the parameters.
The randomness of the parameters is formalized via a prior distribution over the parameter space, and the specified prior distribution is updated via Bayesian conditioning on the data which leads to the so-called posterior distribution (\cite{wasserman2013all, samaniego2010comparison}).
Since the Bayesian prior distribution reflects the (uncertain) beliefs of the modeler about the parameters, the posterior distribution can be considered as updated beliefs about the parameters, which is interpreted as UQ.

This means that \textbf{frequentist statistics has valid behavioral properties}, while \textbf{Bayesian has valid belief updating}.
The two approaches complement each other, however due to the focus on statistical analysis that does not take into account theoretical populations (the dataset is fixed) and easy to interpret UQ that is based on beliefs instead of long run guarantees, Bayesian statistics has been significantly more appealing in ML and still is today (\cite{mackay1995bayesian, theodoridis2015machine, tipping2003bayesian, theodoridis2015machine, murphy2022probabilistic, kristiadi2020being}). 
This has led to the development of algorithms that sample from the posterior distribution efficiently (\cite{chandra2021bayesian, chandra2024bayesian, vehtari2000mcmc, wiese2023towards}).
Although these are valid posterior distributions, the over-parametrized nature of most modern ML applications, makes it unclear how to choose prior distributions, and more importantly \textbf{how to interpret a posterior over an over-parametrized parameter space}.
\citet{wenzel2020good} question the performance of Bayesian Neural Networks on the basis of the performance on new data, by showing that the predictive performance improves when overcounting the evidence, which violates the information processing optimality of the Bayesian updating (\cite{zellner1988optimal}).
This can be justified via the safe Bayesian paradigm under the assumption that the model class is misspecified (\cite{grunwald2011safe}).

In this paper we take a frequentist perspective on UQ, as such our aim is to
\begin{itemize}
\item reinterpret Bayesian updating as an ensemble learning paradigm,
\item question the validity of the predictive distribution by taking a closer look at its long run properties on a population,
\item provide measures of the quality of the Bayesian inferential stage by looking at the frequentist validity of the predictive distribution as a function of the prior,
\item suggest an algorithm to calibrate the predictive distribution that leads to valid frequentist guarantees over the predictions with high probability.
\end{itemize}

Our position is summarized as follows:

\noindent\fbox{%
    \parbox{\textwidth}{%
    \textbf{We question the usefulness of Bayesian UQ in terms of beliefs over the parameters. Under this view, Bayesian posteriors do not meaningfully encode uncertainty, consequently we propose to assess the validity of both the prior and the posterior distribution via frequentist means. As such, we treat the posterior only as a weighing of ensemble of models.
    }%
}
}

\paragraph{Learning task and notation} In this paper we consider a supervised learning scenario. For that purpose a finite training dataset $(X_{i}, Y_{i})_{i=1}^{n} \in (\mathcal{X} \times \mathcal{Y})^{n}$ i.i.d. from some probability measure is available, with an input and output space $\mathcal{X}$ and $\mathcal{Y}$, respectively.
$Y_{i}$ is called a label; in the regression context typically a scalar, in the classification a $K$-dimensional vector for a $K$-class classification problem. $X_{i}$ is a $d$-dimensional input; we treat the input as a vector of fixed dimension, however generalization to other data structures (e.g. matrix, tensor, graph) is possible.
Depending on the learning scenario, a loss function $l: \mathcal{Y} \times \mathcal{Y} \rightarrow \mathbb{R}$ and a function class $\mathcal{F}:=\{f: \mathcal{X} \rightarrow \mathcal{Y}, \text{ where f satisfies certain properties}\}$ are prespecified.
The function class does not have to be parametric (e.g. all measurable functions are considered for the learning of k-Nearest Neighbour), but for many practical applications the class is parametrized (e.g. $f: \mathbb{R}^d \rightarrow \mathbb{R}$, where $f$ is a linear map, or all neural networks with a particular architecture).
The learning task entails identifying an optimal function $f^\star \in \mathcal{F}$, or an optimal ensemble of functions $\mathcal{S}(\mathcal{F})$, where $\mathcal{S}$ is a probability measure over the function class.
The optimal function or ensemble depend on the probability measure over the data. Since the probability measure over the data is generally unknown, one way how to select a single function is via risk minimization over the available dataset $(X_{i}, Y_{i})_{i=1}^{n}$, while for ensembles one solves an additional reweighing optimization problem over the model class under consideration.
We denote by $(X^t, Y^t)$ a training set, by $(X^v, Y^v)$ a quantile estimation set (will become clear later) and by $(X^\text{test}, Y^\text{test})$ a test set.
Furthermore, we denote by $\mathbb{P}$ an abstract probability measure defined on some general measurable space $(\Omega, \sigma(\Omega))$, by $\pi_{0}$ and $p$ a prior distribution over $\mathcal{F}$ and a posterior distribution over $\mathcal{F}$, respectively, and by $P$ a predictive distribution.
In this paper, we explicitly differ between point estimators and random estimators. A point estimator is defined as $T: (\mathcal{X} \times \mathcal{Y})^{n} \rightarrow \mathcal{F}$, while a random estimator is  defined as $T': (\mathcal{X} \times \mathcal{Y})^{n} \rightarrow \mathcal{P}(\mathcal{F})$, where $\mathcal{P}(\mathcal{F})$ is the space of all probability measures over $\mathcal{F}$. 
This means that a point estimator maps from the dataset space $(\mathcal{X} \times \mathcal{Y})^{n}$ to the space of functions, while a random estimator maps to the space of probability measures over $\mathcal{F}$.
In other words, the former chooses the best possible function with respect to the observed data, while the latter chooses the best possible probability measure over the function space with respect to the observed data.

\section{Challenges of Bayesian and Frequentist Procedures}
\paragraph{When data speaks for itself}
As already discussed in the introduction, Bayesian and frequentist statistics make different assumptions about the nature of the data and the parameter space, which is reflected on the inference process. 
However, according to the Bernstein-von Mises theorem (\citet{van2000asymptotic}), for finite-dimensional parameter spaces, under the assumption that the true parameters $\theta_{0}$ of a parametric model are unknown, but fixed, the Bayesian posterior converges (under weak assumptions) asymptotically to a (multivariate) Gaussian centered at the maximum likelihood estimate with a covariance matrix $I(\theta_{0})^{-1}$, where $I(\theta_{0})$ is the Fisher Information evaluated at the true parameters $\theta_{0}$.
This means that Bayesian procedures are able to recover the true model in a frequentist sense.
Such a result implies that for finite-dimensional parameter spaces, asymptotically the data determine the inferential process, which decreases the importance of the prior distribution, and the Bayesian credible intervals have valid frequentist coverage.
Although many ML problems are defined on finite-dimensional parameter spaces, the size of the dataset is of similar order as the size of the parameter space at best, and much smaller in modern ML scenarios, which makes the applicability of the Bernstein-von Mises theorem limited, if not void. As such, this provides a particularly challenging situation for Bayesian UQ of overparametrized models, since on one hand the importance of the prior distribution is high because in the ML regimes there is not enough data to counter its effects on the posterior distribution, and on the other hand, there is no clear guidance on how to choose the prior distribution.
Furthermore, $I(\theta_{0})$ is a function of the parameters of the data generating process, which is often violated for ML tasks because one is interested in optimal prediction, not modeling the true data generating process, and to make matters complicated, a true data generating process might not even exist (\cite{grunwald2019minimum}).

\paragraph{Frequentist pitfalls} Frequentism also faces challenges when performing UQ of modern ML problems because the estimators are already difficult to analyze in an asymptotic sense, but in the finite data regime the analysis becomes even more convoluted (\cite{wakefield2013frequentist}).
In some cases the theoretical analysis of the estimators can be circumvented when bootstrapping is applied (\cite{hesterberg2011bootstrap}).
However, in the case of modern ML issues related to the long training procedures and low amounts of data lead to inadequate approximation of the distribution of the estimator, hence bootstrapping is not generally applicable (\cite{nixon2020bootstrapped}) Furthermore, a frequentist UQ procedure is valid under the assumption that the correct model class has been specified, which as already discussed is difficult to satisfy in the ML context.

\paragraph{A way forward}Given the issues that arise in the context of modern ML, we propose an examination of the validity of Bayesian priors from a frequentist perspective and construction of Bayesian predictive intervals that are valid in a frequentist sense with high probability. As such, this can be seen as adequately strengthening Bayesian inference for finite data regimes, but at the same time weakening the notion of long run validity. 

\section{Bayesian Updating as an Optimization Problem}
A series of papers (\cite{mcallester1998some}, \cite{mcallester2003pac}, \cite{mcallester2003simplified}) shows optimal theoretical guarantees for randomized estimators with respect to the size of the gap between the training and the test error (i.e. generalization gap), which led to the development of the Probably Approximately Correct (PAC)-Bayes framework. 
Since in these papers a general loss function is discussed instead of the standard Bayesian updating, this can be seen as a first step towards relaxation of the Bayesian updating components.
Later on, other work explored random estimators in more detail with respect to the size of the parameter space, the choice of loss function, down and up weighing the likelihood function, concentration of the measure, generalization via entropification (\cite{grunwald2011safe, guedj2019primer, alquier2020concentration, grunwald2019tight}). 
In a recent paper \citet{knoblauch2022optimization} presents an optimization-centric perspective on Bayesian updating as
\begin{equation}
    min_{p \in \mathcal{P}} \{E_{f \sim p}[l(f(x), y)] + D(p || \pi_{0})\},
    \label{eq: bayesoptimization}
\end{equation}
where $\mathcal{P}$ is the space of probability measures over $\mathcal{F}$, $l$ is a loss function, $\pi_{0}$ is a prior probability measure over $\mathcal{F}$, and $D$ a divergence measure between $p$ and $\pi_{0}$.
When $\mathcal{P}$ consists of all probability measures over $\mathcal{F}$, $l$ is the negative log-likelihood loss and $D$ is the $KL$-divergence, the measure that minimizes the optimization problem is the one that is computed via the Bayes rule (\cite{zellner1988optimal}), which shows that for a particular optimization problem the Bayesian update rule is an optimal information processing device.
Since these three components can be chosen by the modeler, this optimization-centric perspective generalizes the choice of a posterior distribution with respect to the size of the probability measure class, the loss function that is used, and the divergence between $p$ and $\pi_{0}$.
As such, this can be seen as a regularized optimization over probability measures with the divergence $D$ treated as the regularizer and the prior distribution as the ''centroid''.
When the divergence component is removed, the optimization boils down to standard point estimation under empirical risk minimization.
This means that the optimization problem defined in equation \ref{eq: bayesoptimization} can be seen as a meta-estimator that encompasses all typical estimation strategies.

\paragraph{Optimization does not give UQ}
Although Bayesian updating is a special case of the optimization problem described in equation \ref{eq: bayesoptimization}, and the posterior can be seen as a random estimator, it does not have any UQ attached to it explicitly, but only minimizes the expected negative log likelihood around the prior distribution.
As such, this means that the validity of any uncertainty statement that is based on the posterior distribution is questionable.
Committing to such a rejection of the interpretation of Bayesian updating as providing UQ, we reinterpret Bayesian posteriors as optimal (wrt. the prior) ensembles of models that are a function of the observed dataset.
This implies that the confidence interval statements that Bayesian posteriors produce can only be of the form ''For an ensemble with prior $\pi_{0}$, $(1-\alpha)\cdot 100 \%$ of all ensemble members as functions of the observed data are in some closed neighborhood $C$'', for $\alpha \in (0, 1)$.
The predictive distribution for new inputs $X^\star = x^\star$ will be valid for statements such as ''For a given ensemble class $(1-\alpha)\cdot 100 \%$ of all outputs for an input $x^\star$ are in some closed neighborhood $C$'', since the predictive distribution depends on the weighted ensemble of the models in $\mathcal{F}$.

\paragraph {Is point estimation conditioning?}The optimization problem defined in equation \ref{eq: bayesoptimization} does not explicitly define probabilistic conditioning; however, as we have seen, for specific choices of loss, divergence and collection of measures, the problem can be seen as Bayesian updating which is indeed probabilistic conditioning.
This opens the question under which conditions can the empirical risk minimizer based optimization problem be considered a case of conditioning, and under which it can be considered just a ''function of the data''?
While our generalization above suggests to look at such a general question, we do not attempt to discuss this issue further but concentrate on the issue of interpreting Bayesian UQ in a frequentist way.

\section{Uncertainty Quantification of Bayesian Ensembles}
When interpreting Bayesian updating as a solution to an ensemble optimization problem, it is still necessary to provide UQ about the ensemble.
Since in this paper we take a strictly frequentist interpretation of UQ, we consider an estimator (point or random) to have valid $(1-\alpha)\cdot100\%$ coverage if the uncertainty region (typically for real valued random variables an interval) contains the true quantity with probability at least $1-\alpha$.
We present three measures to assess the validity of a predictive distribution as a function of the prior $\pi_{0}$ and provide a simple learning algorithm that obtains valid frequentist confidence intervals from a Bayesian posterior with high probability.

\subsection{Quality of Bayesian Priors}
The specification of a prior distribution, together with the divergence measure, and loss function, is one of the key components necessary to optimize the problem in equation \ref{eq: bayesoptimization} for a fixed dataset. As previously discussed, in the context of modern ML the prior distribution has significant impact on the posterior, hence the frequentist validity via Bernstein-von Mises theorem is not guaranteed. Therefore, it is necessary to calibrate the uncertainty regions generated via Bayesian posterior or predictive distributions.

Assuming we commit to a divergence measure and a loss function, one can judge the frequentist quality $Q(\pi_{0})$ of a specific prior distribution $\pi_{0}$ in a frequentist sense as
\begin{equation}
    Q(\pi_{0}) := \mathbb{E}_{X^{t}, Y^{t}}\mathbb{E}_{Y, X^{\star}}[\mathbbm{1}_{K}(Y)],
    \label{eq:quality}
\end{equation}
where 
\begin{equation}
    K:=([c_1, c_2] \textnormal{ s.t. } P(c_1 \leq Y^{\star}\leq c_2 | X^{\star} = x^{\star}, X^{t} = x^{t}, Y^{t} = y^{t}) = 1-\alpha).
    \label{eq: K definition}
\end{equation}
The predictive distribution $P(Y^\star | X^\star, X^t, Y^t)$ is a function of the posterior distribution, which depends on the prior; taking expectation over new data $(X^\star, Y)$ and training datasets $(X^t, Y^t)$, the only remaining object $Q(\pi_{0})$ depends on is the chosen prior $\pi_{0}$.
Equation \ref{eq:quality} can be interpreted as a measure of how misleading the results of the Bayesian UQ (in this case formalized as prediction intervals) are on average.
If the constructed intervals claim to cover $Y$ with probability at least $1-\alpha$, then $Q(\pi_{0})$ should be at least $1-\alpha$.
$Q(\pi_{0})$ is frequentist in nature because it relies on computation of the optimal ensemble for different datasets of the population.
Therefore, this can be seen as a frequentist assessment of a Bayesian prior.

For cases where one cannot afford the presence of a certainty below a threshold, a more pesimisstic quantity is defined as
\begin{equation}
    Q'(\pi_{0}):= \text{inf}_{X^t, Y^t}\mathbb{E}_{Y, X^{\star}}[\mathbbm{1}_{K}(Y)],
    \label{eq: worst_case}
\end{equation}
where $K$ is defined as in equation \ref{eq: K definition}, and depends on the chosen prior distribution in the same way.
This can be seen as quality assessment of the prior from a worst case perspective. It is trivial to show that valid worst case coverage implies valid average coverage.

Finally, a probabilistic definition might bring the best of both quantities, as defined in equations \ref{eq:quality} and \ref{eq: worst_case}, respectively, as

\begin{equation}
    Q''(\pi_{0}):= \mathbb{P}\left((X^t, Y^t) \text{ s.t. }\mathbb{E}_{Y, X^{\star}}[\mathbbm{1}_{K}(Y)]\right) \geq 1-\alpha.
    \label{eq: probabilistic}
\end{equation}
The last definition is probabilistic in nature, since it measures for how many possible datasets the coverage for new data will be valid. If this quantity is close to 1, then it means that most datasets will show regularity, hence will lead to reasonable coverage. Therefore, equation \ref{eq: probabilistic} can be as a softer version of \ref{eq: worst_case}, however it provides more information than the expectation based equation \ref{eq:quality}.

Although such quantities are intuitive to motivate, it is difficult to compute them in practice exactly, hence data-based approximations are necessary.
Alternatively, relying on maximin results and lower bounds is possible when assumptions about the distribution of the data are made.
Operationalizing prior distributions from the perspective of the quality of the uncertainty makes the choice of a prior distribution not based on beliefs, but based on frequentist performance, which would remove the burden of the choice from the modeler, as such making Bayesian inference an objective inferential framework with an external criterion of validity.
\subsection{Frequentist Guarantees Over Ensemble of Models}\label{section42}
In a series of papers \citet{park2019pac, park2020pac} suggest the use of PAC bounds in order to construct frequentist valid confidence intervals that are learned from data and an already selected $f \in \mathcal{F}$.
The problem of confidence interval selection is treated as binary classification.

Inspired by this, we propose a calibration algorithm to construct frequentist valid predictive intervals, with high probability, from predictive distributions based on an ensemble of models.
We assume that an ensemble of models $p(f)$ is already given and that $Y$ is defined on some uncountable subset of $\mathbb{R}$ for ease of presentation, although the approach can be extended to classification and count data.
Typically the ensemble $p(f)$ will be learned via optimization of equation \ref{eq: bayesoptimization} and some training data will be used.
We omit the explicit dependence on the dataset from the notation for brevity.
Finally, we assume that the support of the data is a subset of the support of the predictive distribution (the two can be the same).

Since the ensemble $p(f)$ is already given, for a fixed input $X^\star = x^\star$ we can compute the predictive distribution as $P(Y^\star|X^\star = x^\star) = \int P(Y^\star|X^\star = x^\star, f) dp(f)$.
The goal of our algorithm is to estimate the required width of the quantiles of the predictive distribution in order to achieve correct frequentist coverage with high probability.
For that purpose we use a quantile estimation dataset $(X^v_{i}, Y^v_{i})_{i=1}^{m}$.
As such, the problem can be interpreted as a binary classification problem, where correct classification for some quantile $q$ is defined as the output $Y_{i}^{v}$ for a fixed input $X_{i}^{v}$ being covered by the interval of the predictive distribution $P(Y_{i}^{v,\star}|X^{v}_{i})$.
The intervals for each predictive distribution conditioned on an observation $X_{i}^{v}$ are defined as 
\begin{equation}
    c_{1}(q, X_{i}^{v}):=\text{sup } \{y:P(Y_{i}^{v,\star}\leq y|X_{i}^{v})\leq q\}
\end{equation}
\begin{equation}
    c_{2}(q, X_{i}^{v}):=\text{inf }\{y:P(Y_{i}^{v,\star}\leq y | X_{i}^{v})\geq 1-q\},
\end{equation}
where $c_{1}(q, X_{i}^{v})$ is the lower bound and $c_{2}(q, X_{i}^{v})$ is the upper bound of the interval $[c_{1}(q, X_{i}^{v}), c_{2}(q, X_{i}^{v})]$.
The loss function we commit to is the $0-1$ loss which is $0$ when $Y_{i}^{v}$ is contained in the interval, and $1$ otherwise.
By slightly abusing the notation of the loss function as defined in the introduction, we denote the loss as $l(X_{i}^v, Y_{i}^v, q):=1-\mathbbm{1}_{K_{i}}(Y_{i}^v)$ where $K_{i}:=[c_{1}(q, X_{i}^{v}), c_{2}(q, X_{i}^{v})]$. This allows us to define an empirical risk
\begin{equation}
    \hat{R}((X^v, Y^v), q):=\frac{1}{m}\sum_{i=1}^{m}l(X_{i}^v, Y_{i}^{v}, q).
    \label{eq: frequentist_bayes}
\end{equation}

Optimizing over $q$ is a one dimensional problem, hence although the loss function is not differentiable, it can be easily optimized via grid search. Since we assume a lower bound at $q$ and an upper bound at $1-q$, $q \in (0, 0.5]$, as in frequentist confidence interval construction, we specify $\alpha \in (0, 1)$ as the probability we are willing to accept an observation to not be contained in the interval. Therefore, the grid search stops once $\hat{R}((X^v, Y^v), q)\leq \alpha$ with $\hat{q}$ denoting the solution to the problem. Due to the definition of the problem, $\hat{q}$ leads to symmetric intervals.

The risk in equation \ref{eq: frequentist_bayes} depends on the predictive distribution $P(Y^\star|X^\star = x^\star)$.
Since we assumed that the support of the data is a subset of the support of the predictive distribution, and there are $m$-many observations in the dataset $(X^v, Y^v)$, a sufficiently wide quantile $q$ can make the empirical risk arbitrarily small.
Note that this assumption does not mean that the statistical model has to be correct, because models can have the same support as the data, and still be misspecified (e.g. we assume a linear relationship between the inputs and the output, however there are variables that should enter the model quadratically). This also does not imply that a true data generating process is assumed.

Using the PAC framework, it is possible to get frequentist guarantees for this classification problem of the following form

\begin{equation}
    \mathbb{P}\left((X^v, Y^v) \text{ s.t. } R((X,Y),\hat{q}) \leq \alpha + C(\epsilon)\right) \geq 1 - \epsilon(n).
\end{equation}
Since $C(\epsilon)$ depends on $\epsilon$ which depends on the sample size $n$, for some $n$ it becomes negligibly small (alternatively, one can replace $\alpha$ by some $\alpha' < \alpha$ to address the generalization gap $C(\epsilon)$). It should be noted that $R((X, Y), \hat{q}):= \mathbb{P}(Y \in [c_{1}(\hat{q}, X), c_{2}(\hat{q}, X)])$, which is a valid confidence interval in a frequentist sense with probability $1-\epsilon$. In a future work it remains to show what is the generalization gap of this problem, however due to the low complexity of the classification problem (only one parameter is estimated), we believe it to be low. It should be noted that the optimization problem can be generalized to a two dimensional problem where $q = \{(q_{1}, q_{2}) \in (0, 1)^2 \text { s.t. } q_{1} \leq q_{2}\}$, which provides more flexibility with respect to the width of the confidence intervals, however more data is needed to achieve low error rates.
\section{Simulation Studies}
In this section we demonstrate the effectiveness of the frequentist guarantees over an ensemble of models.
We explore the effectiveness of the calibration in simple models because the methodology is easy to present in such simple classes of functions.
However, since no explicit assumptions were made in the presentation in section \ref{section42}, there are not any limitations of the utility of the method on more complex model classes.

\subsection{Study 1}
\paragraph{Setup} The data is simulated from a linear regression with $20$ parameters and $30$ observations for both the ensemble fitting $(X^t, Y^t)$ and quantile estimation $(X^v, Y^v)$, while $300$ observations are used in the test set $(X^\text{test}, Y^\text{test})$.
We opt for more test observations in order to get a good approximation of the theoretical risk $R((X, Y), \hat{q})$.
For all three datasets, an input is a vector of the form $\{1, x_{1}\dots x_{19}\}$, where the parameters $\beta = \{\beta_{0}, \beta_{1} \dots \beta_{19}\}$ and the features $\{x_{1}\dots x_{19}\}$ are sampled from a standard normal distribution.
The responses $Y$ are sampled from a Gaussian $\mathcal{N}(\mu = X^{T}\beta, \sigma^2=4 )$.
The ensemble fitting is done with Bayesian updating of a prior over $\beta \sim \mathcal{N}_{20}(\mu = i\cdot \mathbf{1}, \Sigma=2\cdot I_{20})$, where $\mathcal{N}_{20}$ is the $20$-dimensional Gaussian, $\mathbf{1}$ is a $20$-dimensional vector of ones, $I_{20}$ is the identity matrix in $\mathbb{R}^{20 \times 20}$, $i \in \{-10\dots10\}$.
Therefore, we have a conventional Bayesian posterior as an ensemble.
Since we specify $21$ different prior distributions, one for each $i \in \{-10\dots 10\}$, we get $21$ different posteriors.
The extremely small or extremely large values of the prior do not correspond to adequate priors, hence in the particular data regime the priors will have large influence on the posterior distribution over $\beta$.
We assume that the variance of the model is known, hence we do not cast a prior over it.
We choose a desired coverage rate of $1-\alpha = 0.9$.
We compare the performance of the calibrated quantiles, as described in section \ref{section42}, with that of the naive predictive intervals, which are typically constructed as values at quantiles $[\frac{\alpha}{2}, 1-\frac{\alpha}{2}]$ of the predictive distribution $P(Y^\star|X^\star=x^\star)$.
Although exact analytical solutions exist for both the posterior distribution over $\beta$ and the predictive distribution over the quantile estimation dataset, we are approaching the problem numerically, because we want to also stress the potential numerical limitations of the approach.
All results are averaged across $10$ random seeds.

\begin{figure}[!htb]
\minipage{0.5\textwidth}
  \includegraphics[width=\linewidth]{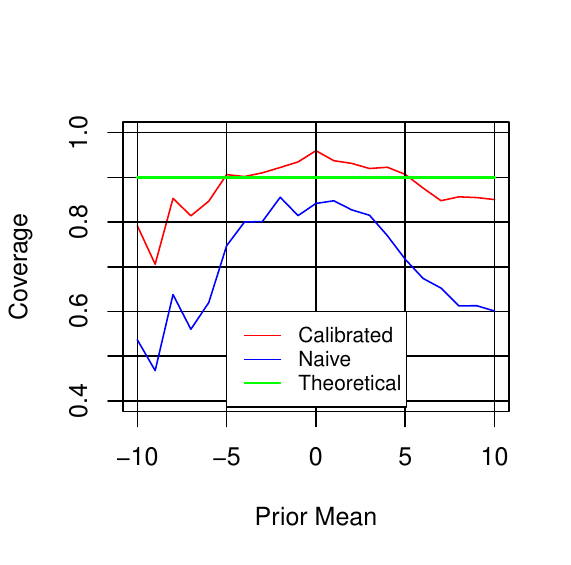}
\endminipage\hfill
\minipage{0.5\textwidth}
  \includegraphics[width=\linewidth]{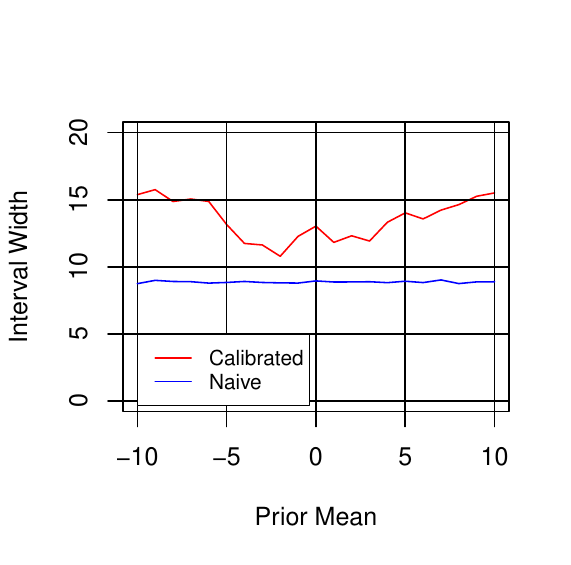}
\endminipage\hfill
\caption{On the left we see coverage of observed test data using the naive approach and the calibrated approach. Each of the values on the $y$ axis are averages over $10$ trials. On the right we see the average width of the predictive intervals for both approaches. The width of the calibrated approach is wider, which is expected since the quantile is chosen on an additional dataset, hence it is informed on the necessary width in order to achieve valid frequentist coverage.}
\label{fig: performance of model}
\end{figure}

\paragraph{Results} In figure \ref{fig: performance of model} we see that the performance of the naive approach never reaches the correct frequentist coverage, and as the prior distribution is far from $0$ the performance drops down to $0.5$, which is quite far from the expected performance.
On the other hand, the calibrated approach reaches the desired performance for priors centered between $-5$ and $5$.
The performance starts to drop below and above $-5$ and $5$, respectively.
This happens because the predictive distribution assigns to the observed data probabilities that are too small to be numerically achievable, hence the samples do not contain such extreme values.
Furthermore, as one would expect, in order to achieve improvement over the coverage, the intervals of the calibrated approach are wider.
This is because the predictive distribution is heavily influenced by the prior, which leads to model predictions that are away from the actual observations, hence in order to adequately address this conflict between the prior and the data, wider intervals are needed in order to achieve correct coverage.
As the conflict between the prior and the data decreases (around 0), the intervals become shorter.
On the other hand, the naive approach is agnostic to the influence of the prior on the predictive intervals, hence the width of the intervals is constant.
\subsection{Study 2}
\paragraph{Setup}The setup of this simulation study is similar to the setup of study 1, with the exception that we introduce an additional feature in the data generation process with parameter $\beta_{20}$.
Since we do not have access to the additional feature, we are fitting a misspecified model.
Because the performance can vary depending on the size of the parameter of the missing variable, we investigate it for parameter values $\beta_{20} = 1$ and $\beta_{20}=3$.
As before, we present both results on the coverage and the width of the predictive intervals.
We fix the probability of coverage to $1-\alpha=0.9$ again as in study 1.

\begin{figure}[!htb]
\minipage{0.5\textwidth}
  \includegraphics[width=\linewidth]{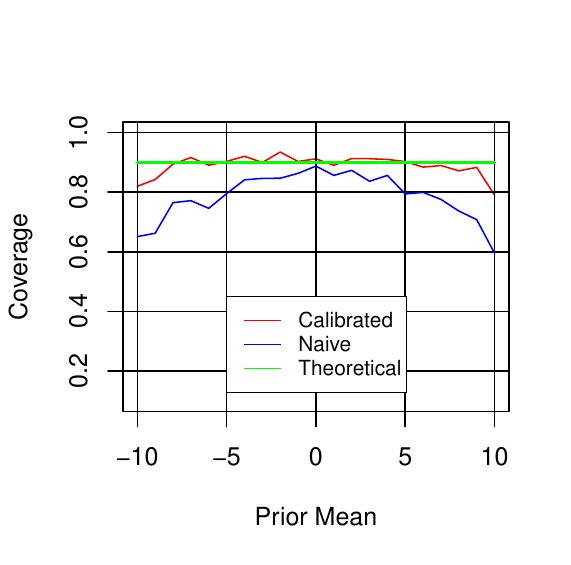}
\endminipage\hfill
\minipage{0.5\textwidth}
  \includegraphics[width=\linewidth]{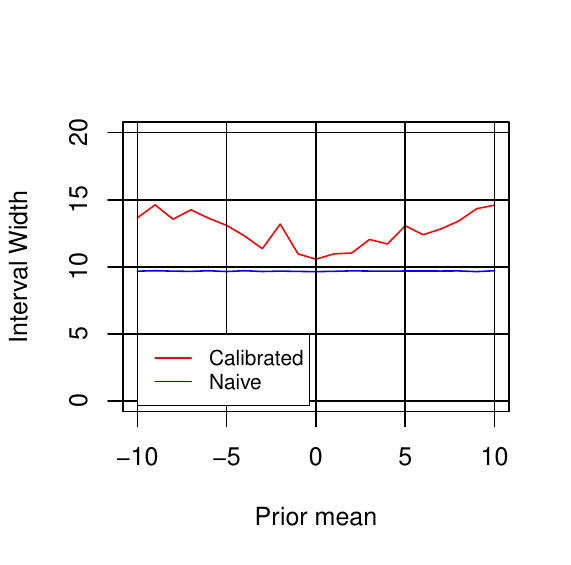}
\endminipage\hfill
\caption{The setup is the same as in figure \ref{fig: performance of model}, the only difference is that in this case there is a missing variable with parameter $\beta_{20}=1$.}
\label{fig: performance of model with missing value 1}
\end{figure}

\begin{figure}[!htb]
\minipage{0.5\textwidth}
  \includegraphics[width=\linewidth]{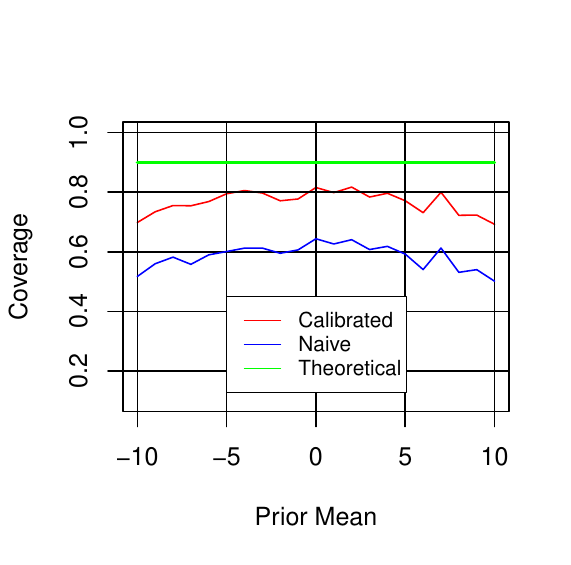}
\endminipage\hfill
\minipage{0.5\textwidth}
  \includegraphics[width=\linewidth]{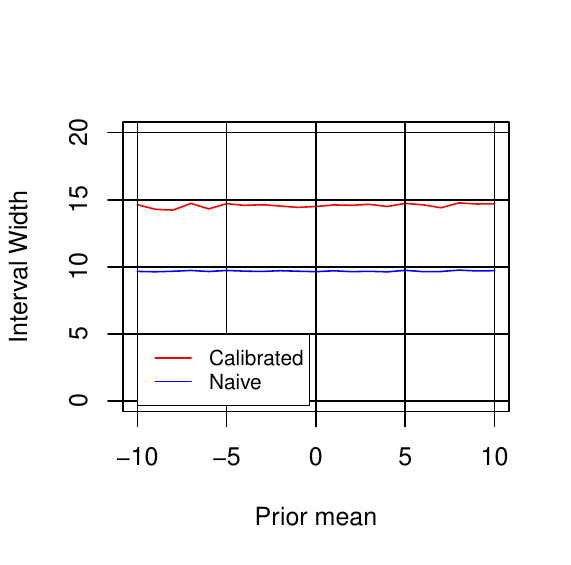}
\endminipage\hfill
\caption{The setup is the same as in figure \ref{fig: performance of model}, the only difference is that in this case there is a missing variable with parameter $\beta_{20}=3$.}
\label{fig: performance of model with missing value 3}
\end{figure}

\paragraph{Results}In figure \ref{fig: performance of model with missing value 1} we observe that similarly as in figure \ref{fig: performance of model} the calibrated approach reaches the desired coverage levels, while the naive approach is unable to account for that.
On figure \ref{fig: performance of model with missing value 3} we observe that both models are unable to achieve the desired frequentist coverage, with the calibrated model achiving better performance.
A major reason why the coverage is not achieved by the calibrated model is because of numerical issues that arise due to the fact that the predictive distribution puts small probability on the region where observations lie, hence it is unable to achieve such a level of numerical precision.
This is a limitation of the approach that we discuss in the next section.

\section{Discussion and Conclusion}
\paragraph{Summary}We argue that since Bayesian updating can be defined as a solution to an optimization problem that does not explicitly consider UQ, it is not justified to treat Bayesian updating as an UQ procedure.
However, since this optimization problem has Bayesian updating as a solution under special conditions, the Bayesian posterior over $\mathcal{F}$ is a valid ensemble.
The components of the optimization problem and its solution can be judged from a frequentist perspective, we focused on two, namely the prior distribution $\pi_{0}(f)$ and the posterior $p(f)$ which consequently leads to the predictive distribution $P(Y^\star|X^\star=x^\star)$ over new observations $X^\star = x^\star$.
We argue that the validity of a prior distribution can be assessed in an average sense in equation \ref{eq:quality}, in a worst case sense in equation \ref{eq: worst_case} and probabilistically in equation \ref{eq: probabilistic}.
Furthermore, we provide a simple algorithm that leads to valid frequentist intervals with high probability that are constructed over the predictive distribution $P(Y^\star|X^\star = x^\star)$ as described in section \ref{section42}.
This suggests that although it is difficult to construct UQ with frequentist validity, and although Bayesian inference does not represent UQ out of the box, it is possible in an additional task to learn valid UQ from an ensemble.
This weakens the frequentist paradigm since now the UQ is not always valid, but only with high probability, however it strengthens the Bayesian position since now we have a precise lower bound on how often we can expect to construct valid UQ.
\paragraph{Limitations} We discuss limitations of both the evaluation of priors and the construction of frequentist valid UQ.
The assessment of the prior distribution as presented in equations \ref{eq:quality}, \ref{eq: worst_case} and \ref{eq: probabilistic} are difficult to compute precisely since the distribution of the data is not generally known.
There are two ways to address this problem, either by lower bounding it when assumptions about the distribution of the data are made, or by taking averages over resampling, which can be computationally expensive.
Depending on the choice of a prior and a statistical model, the grid search optimization of the algorithm described in \ref{section42} over the predictive distribution can be analytically intractable, hence in practical applications the reliance on Markov Chain Monte Carlo sampling might be necessary.
Therefore, although the solution of a well specified problem, as discussed in \ref{section42}, always exists, it is possible to face numerical limitations (e.g. $\hat{q}$ can be of the order $10^{-40}$, which is difficult to reach on a computer) when the model fit is poor or the prior distribution has high influence over the predictive distribution.
\paragraph{Future Work} The work presented in this paper can be extended in multiple directions.
Although this paper presents frequentist measures of the quality of a prior distribution, it does not provide ways to compute these quantities.
Therefore, it is necessary to come up with a methodology, that is statistical model agnostic, to compute exactly or lower bound the quantities.
Furthermore, the generalization error of the calibration algorithm needs to be investigated formally.
Assuming the generalization error is low, it opens up the possibility to check the suitability of statistical models and prior distributions via hypothesis testing of the size of the predictive intervals.

\bibliographystyle{plainnat}
\bibliography{bibliography}
\end{document}